\documentclass{IOS-Book-Article}

\usepackage{mathptmx}
\usepackage[english]{babel}

 \usepackage{graphics}
\usepackage{color}
\usepackage{url}
\usepackage{multirow}

\newcommand{\WORDNET}{WordNet}
\newcommand{\SUMO}{SUMO}

\newcommand{\ADIMENSUMO}{Adimen-SUMO}

\newcommand{\tab}{\hspace{20pt}}
\newcommand{\textConstant}[1]{{\it{#1}}}

\newcommand{\SUMOClassSymbol}{c}
\newcommand{\SUMOIndividualRelationSymbol}{r}
\newcommand{\SUMOIndividualAttributeSymbol}{a}
\newcommand{\SUMOClassOfAttributesSymbol}{A}

\newcommand{\SUMOClass}[1]{\textConstant{#1}$_\SUMOClassSymbol$}
\newcommand{\SUMORelation}[1]{\textConstant{#1}$_\SUMOIndividualRelationSymbol$}
\newcommand{\SUMOAttribute}[1]{\textConstant{#1}$_\SUMOIndividualAttributeSymbol$}
\newcommand{\SUMOClassOfAttributes}[1]{\textConstant{#1}$_\SUMOClassOfAttributesSymbol$}

\newcommand{\synset}[3]{{\textConstant{#1}$_{#3}^{#2}$}}
\newcommand{\WORDNETRelation}[1]{{\it{#1}}}

\newcommand{\equivalenceMappingSymbol}{=}
\newcommand{\subsumptionMappingSymbol}{+}

\newcommand{\equivalenceMapping}[1]{#1$\equivalenceMappingSymbol$}
\newcommand{\subsumptionMapping}[1]{#1$\subsumptionMappingSymbol$}

\def\hb{\hbox to 10.7 cm{}}

\begin{document}

\pagestyle{headings}
\def\thepage{}

\begin{frontmatter}              

\title{Validating \WORDNET{} Meronymy Relations using \ADIMENSUMO{}}

\author[A]{\fnms{Javier} \snm{ \'{A}lvez}%
\thanks{Corresponding Author: Paseo Manuel de Lardiz\'abal, 1 20018 Donostia/San Sebasti\'an. Spain. E-mail: javier.alvez@ehu.eus}},
\author[B]{\fnms{Itziar} \snm{Gonzalez-Dios}}
and
\author[B]{\fnms{German} \snm{Rigau}}

\address[A]{LoRea Group, University of the Basque Country (UPV/EHU)}
\address[B]{IXA Group, University of the Basque Country (UPV/EHU)}

\begin{abstract}
.~In this paper, we report on the practical application of a novel approach for validating the knowledge of \WORDNET{} using \ADIMENSUMO{}. In particular, this paper focuses on cross-checking the \WORDNET{} meronymy relations against the knowledge encoded in \ADIMENSUMO{}. Our validation approach tests a large set of competency questions (CQs), which are derived (semi)-automatically from the knowledge encoded in \WORDNET{}, \SUMO{} and their mapping, by applying efficient first-order logic automated theorem provers. Unfortunately, despite of being created manually, these knowledge resources are not free of errors and discrepancies. In consequence, some of the resulting CQs are not plausible according to the knowledge included in \ADIMENSUMO{}. Thus, first we focus on (semi)-automatically improving the alignment between these knowledge resources, and second, we perform a minimal set of corrections in the ontology. Our aim is to minimize the manual effort required for an extensive validation process. 
%
 We report on the strategies followed, the changes made, the effort needed and its impact when validating the \WORDNET{} meronymy relations using improved versions of the mapping and the ontology. Based on the new results, we discuss the implications of the appropriate corrections and the need of future enhancements.
\end{abstract}

\end{frontmatter}
\markboth{April 2018\hb}{April 2018\hb}

\section{Introduction}


Developing large commonsense knowledge bases such as \WORDNET{} \cite{Fellbaum'98} and \SUMO{} \cite{Niles+Pease'01} is a never-ending task that has been mainly carried out manually. Despite of being created manually, these large knowledge bases are not free of errors and inconsistencies. Fortunately, a few automatic approaches have also been  applied focusing on checking certain structural properties on \WORDNET{} (e.g. \cite{daude2003making} and \cite{richens2008anomalies}) or using automated theorem provers on \SUMO{} (e.g. \cite{HoV06} and \cite{ALR12}). Just a few more have  studied automatic ways to validate the knowledge content encoded in these resources by cross-checking them. For instance, the authors of \cite{AAC08} exploit the EuroWordNet Top Ontology \cite{rodriguez1998top} and its mapping to \WORDNET{} for detecting many ontological conflicts and inconsistencies in the \WORDNET{} nominal hierarchy. 
In \cite{ALR15,ALR17a} we introduced a general framework for automatically cross-checking the knowledge in \WORDNET{} and \SUMO{}, and we proposed  a method for the (semi)-automatic creation of {\it competency questions} (CQs) \cite{GrF95} for evaluating the competency of \SUMO{}-based ontologies like \ADIMENSUMO{} \cite{ALR12}. Our proposal is based on several predefined question patterns (QPs) that are instantiated using information from \WORDNET{} and its mapping into \SUMO{} \cite{Niles+Pease'03}. In addition, we described an application of first-order logic (FOL) automated theorem provers (ATPs) for the automatic evaluation of the proposed CQs. This proposal was used in \cite{AlR18,AGR18} for a preliminary validation of \WORDNET{}, \SUMO{} and their mapping.


\begin{table}[h]
\centering
\begin{tabular}{lllll}
 & \multicolumn{2}{c}{{\bf Part}} & \multicolumn{2}{c}{{\bf Whole}} \\ \hline
{\bf Valid } & \synset{parent}{1}{n} & Subsumed by & \synset{family}{2}{n} & Subsumed by \\
 & a father or mother; (...) & {\bf BiologicalAttribute} &  primary social group; (...) & {\bf FamilyGroup} \\[5pt]  
\hline
{\bf Invalid} & \synset{hyaena}{1}{n} & Subsumed by & \synset{family\_hyaenidae}{1}{n} & Subsumed by \\
 & doglike nocturnal & {\bf Canine} & hyenas & {\bf Canine} \\
 & mammal (...) & & & \\[5pt]
\hline
\end{tabular}
\caption{Valid and invalid examples of the \WORDNETRelation{member} relation}
\label{tab:member-intro-examples}
\end{table}

Nevertheless, the experiments using several state-of-the-art FOL ATPs that are reported in \cite{ALR15,AlR18,AGR18} reveal that a low percentage of the evaluated relation pairs from \WORDNET{} can be currently validated against \ADIMENSUMO{}. We identified three possible causes for this low validation ratio:
\begin{itemize}
\item Discrepancies in the knowledge encoded in \WORDNET{} and \SUMO{}
\item Incorrect mappings between \WORDNET{} and \SUMO{}
\item Limitations of ATPs
\end{itemize}

For instance, in Table \ref{tab:member-intro-examples} we present two examples. The first example is valid because the knowledge from \WORDNET{}, \SUMO{} and its mapping is correctly aligned: individuals with an instance of \textConstant{BiologicalAttribute} as property can be member of instances of \textConstant{FamilyGroup}. The last example is invalid: \textConstant{Canine} is characterized as an individual (i.e. not a group) and, therefore, it cannot have members.


In this paper, we present a (semi)-automatic approach for validating the \WORDNET{} meronymy relations. For this purpose, we apply FOL ATPs on a large set of CQs derived (semi)-automatically from the knowledge encoded in \WORDNET{}, \SUMO{} and their mapping. Unfortunately, these knowledge resources are not free of errors and discrepancies. In consequence, some of the resulting CQs are invalid according to the knowledge encoded in \ADIMENSUMO{}. Thus, we focus on improving the mapping information and, hence, increasing the number of \WORDNET{} relation pairs that can be automatically validated against the knowledge in \ADIMENSUMO{}. 

To this end, our approach consists of three phases. The first two phases involve the mapping between \WORDNET{} and \SUMO{}, and the third one affects \ADIMENSUMO{}.
We describe the changes made, the manual effort required during the correction process and the improvement in the validation of \WORDNET{} meronymy relations; exactly, we spent 26 hours of manual modifications and increased the validation results 35 absolute points due to the improvement achieved in the mapping and the ontology. In addition, we discuss the implication of the performed corrections and the need of future changes.

For example, after the corrections described in this paper, the mapping of the synset \synset{family\_hyaenidae}{1}{n} presented in Table \ref{tab:member-intro-examples} is corrected to be subsumed by \textConstant{GroupOfAnimals}.

{\it Outline of the paper.} In the next section, we describe the knowledge resources and the (semi)-automatic evaluation framework that are used in our work. In Section \ref{sec:Correction} we introduce our approach for correcting both the mapping and the ontology, and we describe its practical application validating the \WORDNET{} meronymy relations. Next, we discuss our experimental results in Section \ref{sec:Discussion}. Finally, we provide some conclusions and discuss future work in Section \ref{sec:Conclussions}.

\section{Knowledge Resources and Evaluation Framework} \label{sec:Framework}

In this section, we describe the knowledge resources that  we have used in our study and define the evaluation framework that enables their automatic validation by means of the use of ATPs.

\WORDNET{} \cite{Fellbaum'98} is a large lexical database of English. Nouns, verbs, adjectives and adverbs are grouped into sets of synonyms (synsets), each one denoting a distinct concept. Synsets are interlinked by means of lexical-semantic relations. \WORDNET{} provides three main meronymy relations that relate noun synsets: i) \WORDNETRelation{part}, the general meronymy relation; ii) \WORDNETRelation{member}, which relates particulars and groups; and iii) \WORDNETRelation{substance}, which relates physical matters and things. In total, \WORDNET{} v3.0 includes 22,187 (ordered) meronymy pairs (around a 10\% of the pairs in \WORDNET{}): 9,097 pairs using \WORDNETRelation{part}, 12,293 pairs using \WORDNETRelation{member} and 797 pairs using \WORDNETRelation{substance}. For example, the synsets \synset{heart\_valve}{1}{n} and \synset{heart}{2}{n} are related by \WORDNETRelation{part}, \synset{lamb}{1}{n} and \synset{genus\_ovis}{1}{n} are related by \WORDNETRelation{member}, and \synset{neuroglia}{1}{n} and \synset{glioma}{1}{n} are related by \WORDNETRelation{substance}.

\ADIMENSUMO{} \cite{ALR12} is a first-order logic (FOL) ontology obtained by means of a suitable transformation of most of the knowledge (around 88\% of the axioms) in the {\it top} and {\it middle} levels of \SUMO{}. \ADIMENSUMO{} enables the application of state-of-the-art FOL  ATPs such as Vampire \cite{KoV13} and E \cite{Sch02} in order to automatically reason on the basis of the knowledge in \SUMO{}. 
We denote the nature of \SUMO{} concepts by adding as subscript the following symbols: $\SUMOIndividualRelationSymbol{}$ for \SUMO{} relations, $\SUMOClassSymbol{}$ for \SUMO{} classes, $\SUMOIndividualAttributeSymbol{}$ for \SUMO{} attributes and $\SUMOClassOfAttributesSymbol{}$ for classes of \SUMO{} attributes, for example: \SUMORelation{material}, \SUMOClass{GroupOfAnimals}, \SUMOAttribute{Solid} and \SUMOClassOfAttributes{BiologicalAttribute}.

Finally, we also exploit the semantic mapping between \WORDNET{} and \SUMO{} \cite{Niles+Pease'03} that connects \WORDNET{} synsets to \SUMO{} concepts. Three semantic relations are used in the mapping between \WORDNET{} and \SUMO: {\it equivalence}, {\it subsumption} and {\it instance}. The mapping relation {\it equivalence} connects \WORDNET{} synsets and \SUMO{} concepts that are semantically equivalent. {\it Subsumption} (or {\it instance}) is used when the semantics of the \WORDNET{} synsets is less general (or instance) than the semantics of the \SUMO{} concepts to which the synsets are connected. For example, the synset \synset{lamb}{1}{n} is connected to \SUMOClass{Lamb} by {\it equivalence} and \synset{neuroglia}{1}{n} is connected to \SUMOClass{Tissue} by {\it subsumption}. From now on, we denote the semantic mapping relations by concatenating the symbols `$\equivalenceMappingSymbol{}$' ({\it equivalence}), `$\subsumptionMappingSymbol${}' ({\it subsumption}) and  `@' ({\it instance}) to the corresponding \SUMO{} concept e.g. \synset{lamb}{1}{n} is connected to \equivalenceMapping{\SUMOClass{Lamb}} and \synset{neuroglia}{1}{n} is connected to \subsumptionMapping{\SUMOClass{Tissue}}.

For the automatic validation of the \WORDNET{} meronymy relations, we apply the evaluation framework introduced in \cite{ALR17a}, which is an adaptation of the method proposed in \cite{GrF95} for the formal design and evaluation of ontologies on the basis of CQs. This framework enables the use of ATPs in order to decide whether a CQ is entailed or not by the ontology. Further, we adapt the method introduced in \cite{AlR18,AGR18} for the automatic creation of CQs on the basis of meronymy by means of four Question Patterns (QPs). Roughly speaking, given a \WORDNET{} meronymy pair, the corresponding QP is selected according to the mapping relation of the related synsets---two options: {\it equivalence} or {\it subsumption}/{\it instance}--- and then instantiated according to the specific \SUMO{} concepts to which synsets are connected. In addition, the \SUMO{} meronymy relation that is used in the resulting CQ is selected according to the \WORDNET{} meronymy relation in the pair. In this paper, we propose to use the following \SUMO{} meronymy relations:
\begin{itemize}
\item \SUMORelation{properPart} if the synsets are related by \WORDNETRelation{part}.
\item \SUMORelation{member} if the synsets are related by \WORDNETRelation{member}.
\item \SUMORelation{material} if the synsets are related by \WORDNETRelation{substance}.
\end{itemize}
It is worth noting that in \cite{AGR18} we use \SUMORelation{part} instead of \SUMORelation{properPart} in order to instantiate QPs for \WORDNET{} \WORDNETRelation{part} pairs. Currently, we think that the semantics of \SUMORelation{properPart} is more similar to the semantics of the \WORDNET{} relation \WORDNETRelation{part} since \SUMORelation{part} is defined as reflexive in \SUMO{}. For example, the synsets \synset{heart\_valve}{1}{n} and \synset{heart}{2}{n} are respectively connected to \subsumptionMapping{\SUMOClass{BodyPart}} and \subsumptionMapping{\SUMOClass{Heart}}. Since the synsets are related by \WORDNETRelation{part} and connected using the mapping relation {\it subsumption}, we use \SUMORelation{properPart} for the instantiation of the first QP proposed in \cite{AlR18} and obtain the following CQ:
\begin{tabbing}
\tab \tab \tab \= (exists (?X ?Y) \\
\> \tab \= (and \\
\> \> \tab \= (\$instance ?X BodyPart) \\
\> \> \> (\$instance ?Y Heart) \\
\> \> \> (properPart ?X ?Y) \\
\> \> ) \\
\> )
\end{tabbing}

Using ATPs, CQs are decided to be {\it passing} (if proved to be entailed by the ontology), {\it non-passing} (their negations are proved to be entailed by the ontology) and {\it unresolved} (nor the CQs neither their negations are proved to be entailed by the ontology). Based on this automatic classification, \WORDNET{} relation pairs can be decided to be {\it validated}, {\it unvalidated} and {\it unknown}. More specifically, a pair is classified as validated, unvalidated or unknown if the corresponding CQ is passing, non-passing or unresolved respectively. In addition, the \WORDNET{} relation pairs that yield to CQs that violate the \SUMO{} domain restrictions are also classified as unvalidated.


\begin{table}[h]
\centering
\resizebox{\textwidth}{!}{
\begin{tabular}{l|rrrrr|rrr}
\multicolumn{1}{c}{{\bf \WORDNET{} relation}} & \multicolumn{1}{c}{{\bf Total}} & \multicolumn{1}{c}{{\bf V}} & \multicolumn{2}{c}{{\bf U}} & \multicolumn{1}{c}{{\bf ?}} & \multicolumn{1}{c}{{\bf Recall}} & \multicolumn{1}{c}{{\bf Precision}} & \multicolumn{1}{c}{${\mathbf {\mathit F1}}$} \\
\hline
\WORDNETRelation{substance} & 797 & 80 & 660 & \hspace{-12pt} (1) & 57 & 0.10 & 0.10 & 0.10 \\
\WORDNETRelation{member} & 12,293 & 19 & 11,963 & \hspace{-12pt} (24) & 311 & 0.00 & 0.00 & 0.00 \\
\WORDNETRelation{part} & 9,097 & 1,255 & 1,444 & \hspace{-12pt} (72) & 6,398 & 0.14 & 0.46 & 0.21 \\
{\bf Total} & {\bf 22,187} & {\bf 1,354} & {\bf 14,067} & {\bf \hspace{-12pt} (97)} & {\bf 6,766} & {\bf 0.06} & {\bf 0.09} & {\bf 0.07} \\
\hline
\end{tabular}
}
\caption{Evaluation of \WORDNET{} meronymy pairs according to the original mapping}
\label{tab:CurrentMappingResults}
\end{table}

In Table \ref{tab:CurrentMappingResults} we report on the initial results obtained by applying the above described evaluation framework to the original versions of \WORDNET{}, \ADIMENSUMO{} and the mapping between \WORDNET{} and \SUMO{}.\footnote{All the practical experimentations reported in this paper have been performed on a Intel\textregistered~Xeon\textregistered~CPU E5-2640v3@2.60GHz with 2GB of RAM per processor. We have used the ATPs Vampire (versions v2.6, v3.0, v4.1 and v4.2.2) and E (v2.1) with an execution time limit of 600 seconds and memory limit of 2 GB. For each test, we provide to the ATP the corresponding conjecture together with the ontology. We have used the following execution parameters with all the versions of Vampire: {\tt --proof tptp --output\_axiom\_names on --mode casc -t 600 -m 2048}. Regarding E, we have used the following execution parameters: {\tt --auto --proof-object -s --cpu-limit=600 --memory-limit=2048}.} For each \WORDNET{} relation (first column), we provide the number of \WORDNET{} relation pairs (Total column) and the number pairs that are validated/unvalidated/unclassified (V, U and ? columns respectively) by following the proposed evaluation framework. In the case of unvalidated pairs, we also provide between brackets the number of pairs that are evaluated by using ATPs. Finally, in the last 3 columns we provide recall (calculated as the ratio between validated pairs and total pairs), precision (calculated as the ratio between validated pairs and validated+unvalidated pairs) and $F$1 (calculated as the harmonic mean of precision and recall) values that result for each \WORDNET{} meronymy relation. From this table, we can conclude that the results are really poor since a very few \WORDNET{} relation pairs are classified as validated and many pairs are classified as unvalidated, especially in the case of \WORDNETRelation{member} and \WORDNETRelation{substance}. In the case of \WORDNETRelation{part}, most of the  pairs are classified as unknown. That is, the direct application of our evaluation methodology just allows to validate a mere 6\% of the meronymy relations encoded in \WORDNET. Apparently, most of the unvalidated pairs correspond to CQs that violate the \SUMO{} domain restrictions (as shown in the unvalid example of Table \ref{tab:member-intro-examples}). This may be an indication of a large number of errors and inconsistencies in the mapping between \WORDNET{} and \SUMO.

\section{Correction Approach} 
\label{sec:Correction}



With the aim of improving the evaluation results reported in Table \ref{tab:CurrentMappingResults} in a cost-effective way, we  carried out a correction process consisting of three phases:
\begin{enumerate}
\item Structural corrections in the mapping based on the \WORDNET{} hierarchy 
\item Opportunistic corrections in the mapping based on the \WORDNET{} information
\item Opportunistic corrections of the ontology and ontology augmentation 
\end{enumerate}
In the following subsections we describe each of these phases.





\subsection{First Phase: Structural Corrections} 
\label{sec:blcs}

In order to perform structural corrections based on the  \WORDNET{} hierarchy, we decided to inspect the  Basic Level Concepts (BLCs) \cite{izquierdo2007exploring}. BLCs are frequent and salient concepts in \WORDNET{} that try to represent as many concepts as possible (abstract concepts) and as many distinctive features as possible (concrete concepts).  

In \WORDNET{} there are 800 BLCs and we decided to inspect manually and individually 200 of them (25~\% of the sample; those being more frequent and having more descendants in \WORDNET{}) to check if their mapping was correct or not. In order to perform this correction, we used information from \WORDNET{}, Top Concept Ontology (TCO) \cite{AAC08}, \SUMO{} and \SUMO{} documentation. For each BLC and based on that information we decided whether the mapping was correct or not, and proposed a new mapping when it was not considered correct. In this correction phase we tried to make as less changes as possible; so, if the original mapping was acceptable we did not change it although it could be a better choice. This way, we were able to revise and correct when necessary around 20 BLCs per hour. In total, we spend 10 hours revising and correcting these BLCs.

\begin{table}[t]
\centering
\begin{tabular}{p{0.8cm}p{1.3cm}p{2.5cm}p{2.5cm}p{2.9cm}}
{\bf Error} & {\bf Synset} & {\bf Gloss} & {\bf Original mapping} & {\bf Corrected mapping}   \\ \hline
Group& \synset{dicot\_genus}{1}{n} & genus of flowering plants (...) & \subsumptionMapping{\SUMOClass{FloweringPlant}} & \subsumptionMapping{\SUMOClass{Group}} \\ \hline  
Group& \synset{fish\_genus}{1}{n}& any of various genus of fish & \equivalenceMapping{\SUMOClass{Fish}} &  \subsumptionMapping{\SUMOClass{GroupOfAnimals}}  \\ \hline
Punctual& \synset{agency}{1}{n} & an administrative unit of government; (...) & \equivalenceMapping{\SUMOClass{PoliticalOrganization}} & \subsumptionMapping{\SUMOClass{GovernmentOrganization}} \\  
\hline
\end{tabular}
\caption{Examples of mapping corrections of BLCs}
\label{tab:correctedBLS}
\end{table}

Following this approach, we corrected the mapping of 52 BLCs manually (26~\%). Then, we  automatically propagated  the corrected BLC mappings to their hyponyms as long as the hyponym and its BLC were equally mapped in the original mapping. This way, a total of 3,883 mappings were corrected. This manual correction can be classified in two types a) groups that are characterized as individual classes (38 synsets, 73~\%), most of them related to plants and animals, and b) punctual mapping errors (14 synsets, 27~\%). As an example, in Table \ref{tab:correctedBLS} we present three synsets with their original mapping to \SUMO{} and their corrected mapping.

\begin{table}[h]
\centering
\resizebox{\textwidth}{!}{
\begin{tabular}{l|rrrrr|rrr}
\multicolumn{1}{c}{{\bf \WORDNET{} relation}} & \multicolumn{1}{c}{{\bf Total}} & \multicolumn{1}{c}{{\bf V}} & \multicolumn{2}{c}{{\bf U}} & \multicolumn{1}{c}{{\bf ?}} & \multicolumn{1}{c}{{\bf Recall}} & \multicolumn{1}{c}{{\bf Precision}} & \multicolumn{1}{c}{${\mathbf {\mathit F1}}$} \\
\hline
\WORDNETRelation{substance} & 797 & 80 & 661 & \hspace{-12pt} (1) & 56 & 0.10 & 0.11 & 0.10 \\
\WORDNETRelation{member} & 12,293 & 21 & 8,341 & \hspace{-12pt} (24) & 3,391 & 0.00 & 0.00 & 0.00 \\
\WORDNETRelation{part} & 9,097 & 1,253 & 1,444 & \hspace{-12pt} (72) & 6,400 & 0.13 & 0.46 & 0.21 \\
{\bf Total} & {\bf 22,187} & {\bf 1,354} & {\bf 10,446} & {\bf \hspace{-12pt} (97)} & {\bf 10,387} & {\bf 0.06} & {\bf 0.11} & {\bf 0.08} \\
\hline
\end{tabular}
}
\caption{Evaluation of \WORDNET{} meronymy pairs according to the structural correction of the mapping}
\label{tab:StructuralCorrectionResults}
\end{table}

The results obtained after this first correction phase are presented in Table  \ref{tab:StructuralCorrectionResults}.  
In this phase, 1,354 meronymy pairs have been validated, 10,446 pairs remain unvalidated and 10,347 pairs are still unclassified. Although the total amount of meronymy pairs classified as either validated or unvalidated decrease with respect to the starting point (see Table \ref{tab:CurrentMappingResults}) after investing 10 hours correcting the mapping, the $F$1 measure just increases a mere 1~\%. Based on its limited impact and the effort invested in the correction, we decided to move forward to the second phase of our approach.


\subsection{Second Phase: Opportunistic Corrections} 
\label{sec:opportunistic}

At this stage of the correction process we decided to inspect the unclassified pairs according to the meronymy relation and we carried out when necessary a few {\it ad hoc} corrections. To ease the inspection, we grouped the pairs 
according to their mapping to \SUMO{} and we ordered them by frequency.  We used the information from \WORDNET{}, \SUMO{} and \SUMO{} documentation for the analysis. Next, we describe the corrections made in the \WORDNETRelation{substance} and \WORDNETRelation{member} pairs. 

\setlength{\tabcolsep}{4pt}

\begin{table}[!h]
\centering
\resizebox{\textwidth}{!}{
\begin{tabular}{lrlllll}
\multicolumn{1}{c}{\multirow{2}{*}{{\bf Group}}} & \multicolumn{1}{c}{\multirow{2}{*}{{\bf \#}}} & \multicolumn{1}{c}{\multirow{2}{*}{{\bf Role}}} & \multicolumn{1}{c}{{\bf Synset}} & \multicolumn{1}{c}{\multirow{2}{*}{{\bf Gloss}}} & \multicolumn{1}{c}{{\bf Original}} & \multicolumn{1}{c}{{\bf Corrected}} \\
\multirow{2}{*}{} & \multirow{2}{*}{} & \multirow{2}{*}{} & \multicolumn{1}{c}{{\bf example}} & \multirow{2}{*}{} & \multicolumn{1}{c}{{\bf mapping}} & \multicolumn{1}{c}{{\bf mapping}} \\ \hline
Medicines & 41 & Whole & \synset{indocin}{1}{n} &a (...) drug (...) & \textConstant{Biologically-} & \subsumptionMapping{\SUMOClass{Medicine}} \\
 & & & & & \subsumptionMapping{\SUMOClass{ActiveSubstance}} & \\ \hline
Cocktails & 21 & Whole & \synset{bloody\_mary}{2}{n}& a cocktail  (...)  & \subsumptionMapping{\SUMOClass{AlcoholicBeverage}} & \subsumptionMapping{\SUMOClass{PreparedFood}} \\ 
 & & & & & & \subsumptionMapping{\SUMOAttribute{Liquid}} \\ \hline

Wood types & 26 & Part &\synset{larch}{1}{n}  & wood  (...) & \subsumptionMapping{\SUMOClass{AnatomicalStructure}} & \subsumptionMapping{\SUMOClass{Wood}} \\ \hline 
Meal parts & 14 & Whole & \synset{soy\_flour}{1}{n} & meal (...) & \subsumptionMapping{\SUMOClass{FruitOrVegetable}} & \subsumptionMapping{\SUMOClass{PreparedFood}} \\ \hline
Oils &11& Part & \synset{soyabean\_oil}{1}{n} & oil (...)     & \subsumptionMapping{\SUMOClass{OrganicThing}} & \subsumptionMapping{\SUMOClass{Oil}} \\ 
 & & & & & \subsumptionMapping{\SUMOClass{Substance}} & \\ \hline
Seed types & 10 & Part & \synset{cacao\_bean}{1}{n} & seed  (...) & \subsumptionMapping{\SUMOClass{FruitOrVegetable}} & \subsumptionMapping{\SUMOClass{Seed}} \\ \hline
Rocks & 7 & Whole & \synset{amphibolite}{1}{n} & a metamorphic & \subsumptionMapping{\SUMOClass{Mineral}} & \subsumptionMapping{\SUMOClass{Rock}} \\
 & & & & rock (...) & & \subsumptionMapping{\SUMOAttribute{Solid}} \\ \hline
Others & 62 & Both & \synset{mint}{4}{n} & the leaves of & \subsumptionMapping{\SUMOClass{FruitOrVegetable}} & \subsumptionMapping{\SUMOClass{PlantLeaf}} \\
 & & & & a mint plant (...) & & \\
\hline
\end{tabular}
}
\caption{Mapping corrections of the \WORDNETRelation{substance} pairs}
\label{tab:corrected-substance}
\end{table}

\setlength{\tabcolsep}{6pt}

\subsubsection{Substance relation}
\label{sec:material}

The \SUMO{} predicate \SUMORelation{material} relates a substance (part) to an object (whole) in \SUMO{}. 
In order to correct the most frequent \WORDNETRelation{substance} errors (total 194 synsets), we performed two kind of corrections: a) replace the \SUMO{} concept or b) add a new \SUMO{} concept. In Table \ref{tab:corrected-substance}, 
we sum up the corrections performed: the corrected group (Group column), the number of corrected pairs from the group (\# column), the role of the corrected synsets ---part or whole--- (Role column) and an example of the correction with its gloss, original mapping and corrected mapping in the following columns.

\subsubsection{Member relation}
\label{sec:member}

The \SUMO{} predicate \SUMORelation{member} relates an individual object as part of group or collection.  
Apparently, most errors found in this type of meronymy pairs are due to the fact that species, genera, families, orders, etc. (taxonomic biological classification) are not connected to \SUMO{} classes representing groups (group errors as presented in \ref{sec:blcs}). 

In order to correct this type of errors we designed two very simple heuristics: 
\begin{enumerate}
\item If the synset is an hyponym of \synset{group}{1}{n} in \WORDNET{} and is connected to a subclass of \SUMOClass{Animal} in \SUMO{}, then map the synset to \subsumptionMapping{\SUMOClass{GroupOfAnimals}}.
\item If the synset is a hyponym of \synset{group}{1}{n} in \WORDNET{} and is connected to a subclass of \SUMOClass{Plant} in \SUMO{}, then map the synset to \subsumptionMapping{\SUMOClass{Group}}.
\end{enumerate}
It is worth noting that there is no concept for representing groups of plants in \SUMO{}.

\begin{table}[h]
\centering
\resizebox{\textwidth}{!}{
\begin{tabular}{l|rrrrr|rrr}
\multicolumn{1}{c}{{\bf \WORDNET{} relation}} & \multicolumn{1}{c}{{\bf Total}} & \multicolumn{1}{c}{{\bf V}} & \multicolumn{2}{c}{{\bf U}} & \multicolumn{1}{c}{{\bf ?}} & \multicolumn{1}{c}{{\bf Recall}} & \multicolumn{1}{c}{{\bf Precision}} & \multicolumn{1}{c}{${\mathbf {\mathit F1}}$} \\
\hline
\WORDNETRelation{substance} & 797 & 80 & 661 & \hspace{-10pt} (1) & 56 & 0.10 & 0.11 & 0.10 \\
\WORDNETRelation{member} & 12,293 & 23 & 6,760 & \hspace{-10pt} (38) & 5,510 & 0.00 & 0.00 & 0.00 \\
\WORDNETRelation{part} & 9,097 & 1,253 & 1,444 & \hspace{-10pt} (72) & 6,400 & 0.14 & 0.46 & 0.21 \\
{\bf Total} & {\bf 22,187} & {\bf 1,356} & {\bf 8,865} & \hspace{-10pt} {\bf (111)} & {\bf 11,966} & {\bf 0.06} & {\bf 0.13} & {\bf 0.08} \\
\hline
\end{tabular}
}
\caption{Evaluation of \WORDNET{} meronymy pairs according to the opportunistic correction of the mapping}
\label{tab:OpportunisticMappingCorrectionResults}
\end{table}

\subsubsection{Evaluation}

The analysis and correction of the second phase lasted 8 hours and a total of 2,124 synsets were corrected. In fact, this stage was easier than the previous one because the annotator worked with more specific pairs of synsets instead of isolated and more generic BLCs.

The evaluation results after the correction phases so far are presented in Table \ref{tab:OpportunisticMappingCorrectionResults}: 
1,356 meronymy pairs have been validated, 8,865 pairs remain unvalidated  and 11,996 pairs are still unclassified ($F$1 8~\%). As in the previous correction phase (see Table \ref{tab:StructuralCorrectionResults}), the number of unvalidated meronymy pairs decreases without a very low impact in the $F$1 measure. 
 



\subsection{Third Phase: Opportunistic Corrections of the Ontology and Ontology Augmentation} \label{sec:ontology}

The final phase of our correction approach focuses on the ontology. At this stage, our objective was to detect and solve the problems in the ontology that prevent the validation of many pairs where the mapping information is correct. For this purpose, we considered two kinds of interventions: on the one hand, correcting a few errors detected in the ontology; on the other hand, augmenting the ontology with new knowledge by including a few more axioms. Next, we describe the changes performed in the ontology for each relation.

\subsubsection{Substance relation} \label{subsec:SubstanceOntology}

After a manual inspection of the unclassified \WORDNETRelation{substance} pairs, we detected that all of them were referring to organisms. Thus, we manually reviewed the substances related to organisms that were defined in the ontology and detected that the \SUMO{} concepts \SUMOClass{BodySubstance}, \SUMOClass{AnimalSubstance} and \SUMOClass{PlantSubstance} were lacking of a proper characterization. More specifically, no axiom characterizes the semantics of \SUMOClass{BodySubstance}, and \SUMOClass{AnimalSubstance} and \SUMOClass{PlantSubstance} are defined by a single axiom that incorrectly uses the \SUMO{} predicate \SUMORelation{part} instead of \SUMORelation{material}. Similarly, the axiom that characterizes the semantics of \SUMOClass{Bone} incorrectly uses \SUMORelation{part}. Finally, the concepts \SUMOClass{Hormone} and \SUMOClass{Blood} are characterized as subclasses of \SUMOClass{BodySubstance} although these substances are present only in animals.

We corrected all these problems by updating the ontology as follows:

\begin{itemize}
\item We  properly replaced the \SUMO{} predicate \SUMORelation{part} by \SUMORelation{material} in the axioms that respectively characterize \SUMOClass{AnimalSubstance}, \SUMOClass{PlantSubstance} and \SUMOClass{Bone} (3 axioms corrected).
\item We corrected the characterization of \SUMOClass{Hormone} and \SUMOClass{Blood} to be subclass of \SUMOClass{AnimalSubstance} (2 axioms corrected).
\item We  augmented the characterization of \SUMOClass{BodySubstance}, \SUMOClass{AnimalSubstance} and \SUMOClass{PlantSubstance} by stating that any instance of \SUMOClass{Organism}/\SUMOClass{Animal}/\SUMOClass{Plant} is material of some instance of \SUMOClass{BodySubstance}/\SUMOClass{AnimalSubstance}/\SUMOClass{PlantSubstance} respectively (3 new axioms), and {\it vice versa} (3 new axioms).
\end{itemize}
To sum up, 4 axioms have been corrected and 6 new axioms have been included in the ontology with a total manual effort of 2 hours.

\subsubsection{Member relation} \label{subsec:MemberOntology}

As in the case of \WORDNET{} \WORDNETRelation{substance} pairs, most of the unvalidated or unclassified \WORDNETRelation{member} pairs are related to animals and plants.

With respect to unvalidated \WORDNETRelation{member} pairs, the problems are related to the domain restrictions of the \SUMO{} predicate \SUMORelation{member}. In particular, the first argument of \SUMO{} \SUMORelation{member} pairs is restricted to be instance of \SUMOClass{SelfConnectedObject}, which is disjoint with the \SUMO{} class \SUMOClass{Collection} and hence disjoint with the \SUMO{} class \SUMOClass{Group}. Consequently, we cannot construct a \SUMO{} statement that expresses that an instance of \SUMOClass{Group} is member of another instance of \SUMOClass{Group}, as required for the validation of the examples in Table \ref{tab:member-intro-examples}. This problem was corrected by replacing the domain restriction of the first argument of the \SUMO{} predicate \SUMORelation{member}. In our proposal, the first argument of \SUMO{} \SUMORelation{member} pairs is now restricted to be instance of \SUMOClass{Object}, which is superclass of \SUMOClass{Group} (1 axiom corrected). In addition, the characterization of \SUMOClass{GroupOfPeople} and \SUMOClass{GroupOfAnimals} has to be accordingly updated: in the new proposed axiomatization, an instance of \SUMOClass{GroupOfPeople} can be member of instances of either \SUMOClass{Human} or \SUMOClass{GroupOfPeople}, and the members of an instance of \SUMOClass{GroupOfAnimals} can be either instances \SUMOClass{Animal} that are not instance of \SUMOClass{Human} or instances of \SUMOClass{GroupOfAnimals} (2 axioms corrected).

Regarding unclassified \WORDNETRelation{member} pairs, we created and properly characterized a new \SUMO{} class called \SUMOClass{GroupOfPlants} in order to obtain a more precise mapping for the groups of plants (3 new axioms). For this purpose, we  also redefined the second heuristic presented in Subsection \ref{sec:member}: if the synset is an hyponym of the synset \synset{group}{1}{n} in \WORDNET{} and is connected to a subclass of \SUMOClass{Plant} in \SUMO{} then we map the synset to \subsumptionMapping{\SUMOClass{GroupOfPlants}} (225 mappings updated\footnote{In the second correction phase, these synsets were mapped to \subsumptionMapping{\SUMOClass{Group}}.}). Additionally, we augmented the ontology as follows:

\begin{itemize}
\item Any instance of \SUMOClass{Agent} is member of some instance of \SUMOClass{Group} (1 new axiom).
\item Any instance of \SUMOClass{Human} is member of some instance of \SUMOClass{GroupOfPeople}, \SUMOClass{AgeGroup}, \SUMOClass{FamilyGroup}, \SUMOClass{SocialUnit}, \SUMOClass{EthnicGroup} and \SUMOClass{BeliefGroup} (6 new axioms).
\item Any instance of \SUMOClass{Animal} that is not instance of \SUMOClass{Human} is member of some instance of \SUMOClass{GroupOfAnimals} and \SUMOClass{Brood} (2 new axioms).
\end{itemize}

In total, 3 axioms have been corrected and 12 new axioms have been included in the ontology with a total manual effort of 2 hours.

\subsubsection{Part relation} \label{subsec:PartOntology}

By inspecting the ontology, we have detected that most of the unclassified \WORDNET{} \WORDNETRelation{part} pairs are due to the general use of the \SUMO{} predicate \SUMORelation{part} in the ontology, while few axioms were using the \SUMO{} predicate \SUMORelation{properPart}. The use of the \SUMO{} predicate \SUMORelation{part} is not precise enough and yield a too weak axiomatization since \SUMORelation{part} is defined as reflexive: that is, the \SUMO{} predicate \SUMORelation{part} relates any instance of \SUMOClass{Object} with itself. This fact prevents ATPs from proving many conjectures: it is not possible to infer the existence of two different instances related by the \SUMO{} predicate \SUMORelation{properPart} from the existence of two instances related by the \SUMO{} predicate \SUMORelation{part}, because the two instances may be equal in the latter case. A similar problem caused by the general use of the \SUMO{} predicate \SUMORelation{part} is described in \cite{ALR15}.

In order to overcome the above mentioned problem, we corrected the ontology by semi-automatically replacing \SUMORelation{part} with \SUMORelation{properPart} where convenient. In total, 358 axioms were corrected.

In addition, we  augmented the axiomatization of some concepts in the ontology related to the anatomical parts of organisms. More specifically, we  included the following restrictions:
\begin{itemize}
\item Any instance of \SUMOClass{AnatomicalStructure} and \SUMOClass{BodyPart} is proper part of some instance of \SUMOClass{Organism} (2 new axioms), and {\it vice versa} (2 new axioms).
\item Any instance of \SUMOClass{AnimalAnatomicalStructure} and \SUMOClass{Meat} is proper part of some instance of \SUMOClass{Animal} (2 new axioms), and {\it vice versa} (2 new axioms).
\item Any instance of \SUMOClass{PlantAnatomicalStructure} is proper part of some instance of \SUMOClass{Plant} (1 new axiom), and {\it vice versa} (1 new axiom).
\item Any instance of \SUMOClass{AbnormalAnatomicalStructure} is proper part of some instance of \SUMOClass{Organism} (1 new axiom).
\end{itemize}

Summing up, 358 axioms were semi-automatically corrected and 11 new axioms were included in the ontology with a human effort of 4 hours.

\subsubsection{Results}


\begin{table}[h]
\centering
\resizebox{\textwidth}{!}{
\begin{tabular}{l|rrrrr|rrr}
\multicolumn{1}{c}{{\bf \WORDNET{} relation}} & \multicolumn{1}{c}{{\bf Total}} & \multicolumn{1}{c}{{\bf V}} & \multicolumn{2}{c}{{\bf U}} & \multicolumn{1}{c}{{\bf ?}} & \multicolumn{1}{c}{{\bf Recall}} & \multicolumn{1}{c}{{\bf Precision}} & \multicolumn{1}{c}{${\mathbf {\mathit F1}}$} \\
\hline
\WORDNETRelation{substance} & 797 & 81 & 661 & (1) & 55 & 0.10 & 0.10 & 0.10 \\
\WORDNETRelation{member} & 12,293 & 4,167 & 802 & (51) & 7,324 & 0.34 & 0.84 & 0.48 \\
\WORDNETRelation{part} & 9,097 & 2,436 & 1,379 & (7) & 5,282 & 0.27 & 0.64 & 0.38 \\
{\bf Total} & {\bf 22,187} & {\bf 6,684} & {\bf 2,842} & {\bf (59)} & {\bf 12,661} & {\bf 0.30} & {\bf 0.70} & {\bf 0.42} \\
\hline
\end{tabular}
}
\caption{Evaluation of \WORDNET{} meronymy pairs according to the opportunistic correction of the ontology}
\label{tab:OpportunisticOntologyCorrectionResults}
\end{table}

The results of the previous correction phases are presented in Table \ref{tab:OpportunisticOntologyCorrectionResults}. This time, 6,684 meronymy pairs are validated, 2,842 pairs remain unvalidated and 12,661 pairs are still unclassified ($F$1 42~\%). It seems that making small changes in the ontology brings a large improvement in the results. Additionally, looking at the recall (30~\%), we are validating around one-third of the \WORDNET{} meronymy knowledge, 24 points more than in the initial state (see Table \ref{tab:CurrentMappingResults}).

\begin{table}[!ht]
\centering
\begin{tabular}{l|rrrr}
\multicolumn{1}{c}{\multirow{2}{*}{\textbf{Correction Phase}}} & \multicolumn{1}{c}{\multirow{2}{*}{\textbf{Hours}}} & \multicolumn{2}{c}{\textbf{Corrections}} &  \multicolumn{1}{c}{\multirow{2}{*}{\textbf{Improvement}}} \\
\multicolumn{1}{c}{\multirow{2}{*}{}} & \multirow{2}{*}{} & \textbf{Mappings} & \textbf{Axioms} & \multirow{2}{*}{} \\
\hline
First phase    & 10     & 3,883 & - & $+$1 \\
Second phase  & 8   & 2,124 & -  &      $+$0                   \\
Third phase & 8 & (225) & 369 & $+$34   \\
 \textbf{Total}     & \textbf{26}      & \textbf{6,007}  & \textbf{369}  &     \textbf{$+$35}     \\
 \hline 
 \end{tabular}
 \caption{Summary of the correction process}
 \label{tab:esfuerzo}
\end{table}

\section{Discussion} \label{sec:Discussion}

In Table \ref{tab:esfuerzo} we sum up the manual effort invested only in the correction process\footnote{We just count the manual annotation effort. We are not counting the time of discussions, analysis of the results, etc.}, the number of synsets and axioms corrected and the improvements in the $F$1 measure. Summing up, spending  26 hours of correction brings to an improvement of 35 points in $F$1.   


\begin{table}[!ht]
\centering
\resizebox{\textwidth}{!}{
\begin{tabular}{l|rrrrr|rrr}
\multicolumn{1}{c}{{\bf \WORDNET{} relation}} & \multicolumn{1}{c}{{\bf Total}} & \multicolumn{1}{c}{{\bf V}} & \multicolumn{2}{c}{{\bf U}} & \multicolumn{1}{c}{{\bf ?}} & \multicolumn{1}{c}{{\bf Recall}} & \multicolumn{1}{c}{{\bf Precision}} & \multicolumn{1}{c}{${\mathbf {\mathit F1}}$} \\
\hline
\WORDNETRelation{substance} & 797 & 85 & 659 & (0) & 53 & 0.11 & 0.11 & 0.11 \\
\WORDNETRelation{member} & 12,293 & 95 & 11,964 & (25) & 234 & 0.01 & 0.01 & 0.01 \\
\WORDNETRelation{part} & 9,097 & 2,590 & 2,245 & (873) & 4,262 & 0.28 & 0.54 & 0.37 \\
{\bf Total} & {\bf 22,187} & {\bf 2,770} & {\bf 14,868} & {\bf (898)} & {\bf 4,549} & {\bf 0.12} & {\bf 0.17} & {\bf 0.14} \\
\hline
\end{tabular}
}
\caption{Evaluation of \WORDNET{} meronymy pairs according to the original mapping and opportunistic correction of the ontology}
\label{tab:OriginalMappingAugmentedOntology}
\end{table}

It is easy to see from the results reported in Tables \ref{tab:CurrentMappingResults}-\ref{tab:OpportunisticOntologyCorrectionResults} that we have substantially improved the evaluation results of the meronymy knowledge in \ADIMENSUMO{} and its mapping to \WORDNET{} by means of a very limited number of  corrections. For instance, the $F$1 measure has raised from 0.07 to 0.42. However, being incremental, from the results in those tables it is not possible to infer the contribution of the corrections in each particular phase. For this reason, we have completed our experimentation by performing a new evaluation using the original mapping and the ontology obtained from our last correction phase. The results of this last experimentation are reported in Table \ref{tab:OriginalMappingAugmentedOntology}. These results lead us to two main conclusions: on the one hand, the impact of the corrections in the third phase (correcting the ontology) is greater than the impact of the corrections in the two first phases (correcting the mapping); on the other hand, the improvement is much greater when combining the corrections performed along the three phases. 

\begin{table}[!ht]
\centering
\begin{tabular}{l|rrr|r}
\multicolumn{1}{c}{{\bf Correction Phase}} & \multicolumn{1}{c}{{\bf Passing}} & \multicolumn{1}{c}{{\bf Non-passing}} & \multicolumn{1}{c}{{\bf Unresolved}} & \multicolumn{1}{c}{{\bf Total}} \\
\hline
Initial evaluation & 104 & 28 & 2,088 & {\bf 2,140} \\
First phase & 103 & 28 & 2,054 & {\bf 2,185} \\
Second phase & 104 & 36 & 2,137 & {\bf 2,277} \\
Third phase & 499 & 27 & 2,382 & {\bf 2,382} \\
\hline
\end{tabular}
\caption{Competency Questions used for the evaluation of \WORDNET{} meronymy}
\label{tab:EvaluationOfCQs}
\end{table}

Another indicator that reflects the impact of our proposed corrections is the number of CQs that are obtained from the meronymy QPs and, among them, the number of CQs that can be solved by the ATPs. In Table \ref{tab:EvaluationOfCQs}, we report on the number of CQs that are obtained in the initial evaluation of \WORDNET{} meronymy and after each correction phase (Total Column). In addition, we also present the number of passing, non-passing and unresolved CQs (Passing, Non-passing and Unresolved columns respectively). 

The results reported in Table \ref{tab:EvaluationOfCQs} shows  that both the total number of  CQs and the number of passing CQs always increases after each correction phase, which confirms the improvement on the competency of the meronymic knowledge of \ADIMENSUMO{}. This way, we are able to validate more  meronymy  pairs of \WORDNET{}. 

Regarding the number of non-passing CQs, our results are not significant due to two main reasons: on the one hand, the variations are really small compared to the total number of CQs; on the other hand, we have experimentally checked that ATPs run out of resources when trying to prove the negation of some CQs that are entailed by \ADIMENSUMO{}. More concretely, we have verified that all CQs classified as  non-passing  after the second correction phase will be non-passing after the third correction phase  as well if the execution time limit is enlarged.





Additionally, we have performed an error analysis selecting a subset of most representative CQs. That is, those CQs obtained from at least three unvalidated or unclassified meronymy pairs. A first inspection reveals:
\begin{itemize}
\item Lack of additional meronymic knowledge about organisms in the ontology.
\item Lack of proper metonymy characterization. Some groups should inherit attributes that are characteristic of their individual members. For instance, mappings where individual attributes are applied to groups of people related to professions, positions and social roles e.g. an \synset{economic\_expert}{1}{n} (mapped to \subsumptionMapping{\SUMOClassOfAttributes{Profession}})  can be member of the \synset{economics\_profession}{1}{n} (mapped to \subsumptionMapping{\SUMOClassOfAttributes{SocialRole}}).
\item Ontological decisions. For instance, the axiomatization of \SUMOClass{GroupOfAnimals} in \SUMO{} excludes groups of human beings. As in \SUMO{} \SUMOClass{Human} is subclass of \SUMOClass{Animal}, we cannot validate examples such as \synset{homo\_sapiens}{1}{n} (mapped to \subsumptionMapping{\SUMOClass{Human}}) is member of the \synset{genus\_homo}{1}{n} (mapped to  \subsumptionMapping{\SUMOClass{GroupOfAnimals}}). 
\item Lack of resources for ATPs. Many \WORDNETRelation{member} pairs relating animals have not been validated. For example, pairs having synsets mapped to \subsumptionMapping{\SUMOClass{Fish}} and \subsumptionMapping{\SUMOClass{GroupOfAnimals}} respectively. In the same way, many \WORDNETRelation{part} pairs relating buildings, cities, rivers, etc. and their respective cities, regions, countries $\ldots$  are not validated. We have experimentally checked that this kind of unclassified pairs can be validated by enlarging the execution time limit provided to ATPs. 
\end{itemize}


Finally, we would like to point out the long tail problem, since most of the errors we found just affect a very limited number of \WORDNET{} meronymy relations. In fact, there are many punctual mapping errors. Additionally, the mapping of four synsets was not automatically corrected during the second phase as we were too conservative.

\section{Conclusion and Future Work} \label{sec:Conclussions}

In this paper we reported on the practical application of a novel approach for validating the meronymic knowledge of \WORDNET{} using \ADIMENSUMO{}. To this end, we applied FOL ATPs on a large set of CQs derived (semi)-automatically from the knowledge encoded in \WORDNET{}, \SUMO{} and their mapping.  Trying to minimize the manual effort involved, we focused on improving the mapping information and, hence, increasing the number of \WORDNET{} relation pairs that can be automatically validated against the knowledge in \ADIMENSUMO{}.  

An important result of our research is that it seems to be worth investing effort correcting the ontology. Changes that we carried out at that stage have a major impact in the results, although the structural and the opportunistic corrections also contribute to the improvement. Indeed, just investing a total amount of 26 manual correction hours, we improve the $F$1 results 35 absolute points. Before improving both the mapping and the ontology, we validated just a mere 6~\% of the \WORDNET{} meronymy relations. After applying our (semi)-automatic approach, we are able to validate around one-third of the \WORDNET{} meronymy relations. Moreover, during the correction process, we have also improved both the mapping and the competency of \ADIMENSUMO{}. All the resources ---the corrected mapping, the augmented ontology  and the experimental reports--- are available at \url{http://adimen.si.ehu.es/web/AdimenSUMO}.

We also carried out an initial error analysis that raised aspects for future work. For instance, metonomic and ontological issues (e.g. classifying humans as animals). Moreover, we plan to test if the improved ontology also obtains better results in other benchmarks based on antonymy and semantic roles \cite{ALR17a}. Further, we would like to carry out similar experiments in other datasets (e.g. BLESS\footnote{\url{https://sites.google.com/site/geometricalmodels/shared-evaluation}} \cite{BaL10}) and also validate additional \WORDNET{} semantic relations such as cause and capability. In particular, we intend to concentrate our effort on correcting and augmenting the ontology.

Longer term research  includes a new mapping between \WORDNET{} and \ADIMENSUMO{} on the basis of formulae instead of labels. The aim will be to provide a more precise definition of the semantics of synsets in terms of \ADIMENSUMO{}. 

\bibliographystyle{ieeetr}
\bibliography{references}

\end{document}